\documentclass[sigconf]{acmart}

\AtBeginDocument{%
  \providecommand\BibTeX{{%
    \normalfont B\kern-0.5em{\scshape i\kern-0.25em b}\kern-0.8em\TeX}}}

\setcopyright{acmcopyright}

\copyrightyear{2023}
\acmYear{2023}
\setcopyright{acmlicensed}\acmConference[KDD '23]{Proceedings of the 29th ACM SIGKDD Conference on Knowledge Discovery and Data Mining}{August 6--10, 2023}{Long Beach, CA, USA}
\acmBooktitle{Proceedings of the 29th ACM SIGKDD Conference on Knowledge Discovery and Data Mining (KDD '23), August 6--10, 2023, Long Beach, CA, USA}
\acmPrice{15.00}
\acmDOI{10.1145/3580305.3599525}
\acmISBN{979-8-4007-0103-0/23/08}

\usepackage{multirow,tabularx}
\usepackage{hhline}
\usepackage{amsmath}
\usepackage{bm}
\usepackage{algorithm}
\usepackage{algpseudocode} 
\usepackage{algorithmicx}
\usepackage{subfigure}
\usepackage{color}
\newsavebox{\measurebox}
\usepackage{caption}
\usepackage{graphicx}
\usepackage{makecell}

\newtheorem{problem}{Problem}

\theoremstyle{plain}
\newtheorem{theorem}{Theorem}[section]

\theoremstyle{definition}

\newtheorem{property}[theorem]{Property}
\theoremstyle{remark}
\newtheorem{remark}[theorem]{Remark}

\usepackage{expl3}
\ExplSyntaxOn
\newcommand\latinabbrev[1]{
  \peek_meaning:NTF . {
    #1\@}%
  { \peek_catcode:NTF a {
      #1.\@ }%
    {#1.\@}}}
\ExplSyntaxOff

\def\eg{\latinabbrev{e.g}}

\def\ie{\latinabbrev{i.e}}
\newcommand{\name}{{\textsc{CaliRare}}}

\begin{document}

\title{Towards Reliable Rare Category Analysis on Graphs via Individual Calibration}

\author{Longfeng Wu}
\affiliation{
  \institution{Virginia Tech}
  \city{Blacksburg}
  \state{VA}
  \country{USA}
  \postcode{24060}
}
\email{longfengwu@vt.edu}

\author{Bowen Lei}
\affiliation{
  \institution{Texas A\&M University}
  \city{College Station}
  \state{TX}
  \country{USA}}
\email{bowenlei@stat.tamu.edu}

\author{Dongkuan Xu}
\affiliation{
  \institution{North Carolina State University}
  \city{Raleigh}
  \state{N.C.}
  \country{USA}
}
\email{dxu27@ncsu.edu}

\author{Dawei Zhou}
\affiliation{
 \institution{Virginia Tech}
 \city{Blacksburg}
 \state{VA}
 \country{USA}}
\email{zhoud@vt.edu}

\renewcommand{\shortauthors}{Longfeng Wu, Bowen Lei, Dongkuan Xu, \& Dawei Zhou}
\begin{abstract}
Rare categories abound in a number of real-world networks and play a pivotal role in a variety of high-stakes applications, including financial fraud detection, network intrusion detection, and rare disease diagnosis. Rare category analysis (RCA) refers to the task of detecting, characterizing, and comprehending the behaviors of minority classes in a highly-imbalanced data distribution. 
While the vast majority of existing work on RCA has focused on improving the prediction performance, a few fundamental research questions heretofore have received little attention and are less explored: \emph{How confident or uncertain is a prediction model in rare category analysis? How can we quantify the uncertainty in the learning process and enable reliable rare category analysis? }

To answer these questions, we start by investigating miscalibration in existing RCA methods. Empirical results reveal that state-of-the-art RCA methods are mainly over-confident in predicting minority classes and under-confident in predicting majority classes. Motivated by the observation, we propose a novel individual calibration framework, named \name, for alleviating the unique challenges of RCA, thus enabling reliable rare category analysis. In particular, to quantify the uncertainties in RCA, we develop a node-level uncertainty quantification algorithm to model the overlapping support regions with high uncertainty; to handle the rarity of minority classes in miscalibration calculation, we generalize the distribution-based calibration metric to the instance level and propose the first individual calibration measurement on graphs named Expected Individual Calibration Error (EICE). We perform extensive experimental evaluations on real-world datasets, including rare category characterization and model calibration tasks, which demonstrate the significance of our proposed framework. 

\end{abstract}





\maketitle

\section{Introduction}
In contrast to the massive network data being generated and used every day, it is often the rare occurrences that might be of interest to us and plays a crucial role in various application domains. For example, in financial transaction networks~\cite{bay2006large}, only a small portion of transactions are fraudulent, but it can lead to immeasurable financial loss; 
in network security~\cite{wu2007local}, identifying malicious activities from large amounts of network traffic can better protect users from potential threats; in patient-symptom network~\cite{karim2019comprehensive}, identifying and forecasting rare diseases (\ie, the ones with very few records but severe symptoms) has become a longstanding research problem. Rare category characterization~\cite{mullapudi2021learning} refers to the problem of ``\emph{finding needles from the hay}'', which aims to characterize the support regions of minority classes (\eg, needles) from the overwhelmed majority classes (\eg, hay). 

Tremendous efforts have been made to develop theories and algorithms for analyzing rare categories, from detecting rare categories in the cold-start setting~\cite{pelleg2004active, DBLP:conf/kdd/ZhouZYATDH17} to tracking rare categories in the dynamic system~\cite{ranshous2015anomaly, zhou2017discovering},  
from visualizing the network layout of rare categories~\cite{zhao2014fluxflow, zhang2017survey, lin2017rclens, pan2020rcanalyzer}
to interpreting the behavior of rare categories ~\cite{macha2018explaining, liu2019towards}.
However, the vast majority of existing work does not take into account the reliability of rare category characterization. While many high-stake industries follow highly-regulated processes, prediction models are required to be not only accurate but also reliable. For example, in the case of financial fraud detection, guileful fraudulent users are good at camouflaging themselves in order to bypass the security system owned by financial institutions. Thus, it would be desirable to understand how accurate the prediction models are, as well as how trustworthy they are so that the security systems can raise alarms and request human intervention when predictions are made with low confidence. 

\begin{figure}[ht]
\small
\captionsetup{font=footnotesize}
\centering
\subfigure[Financial transaction network]{
    \includegraphics[width=3.8cm, height=3.1cm]{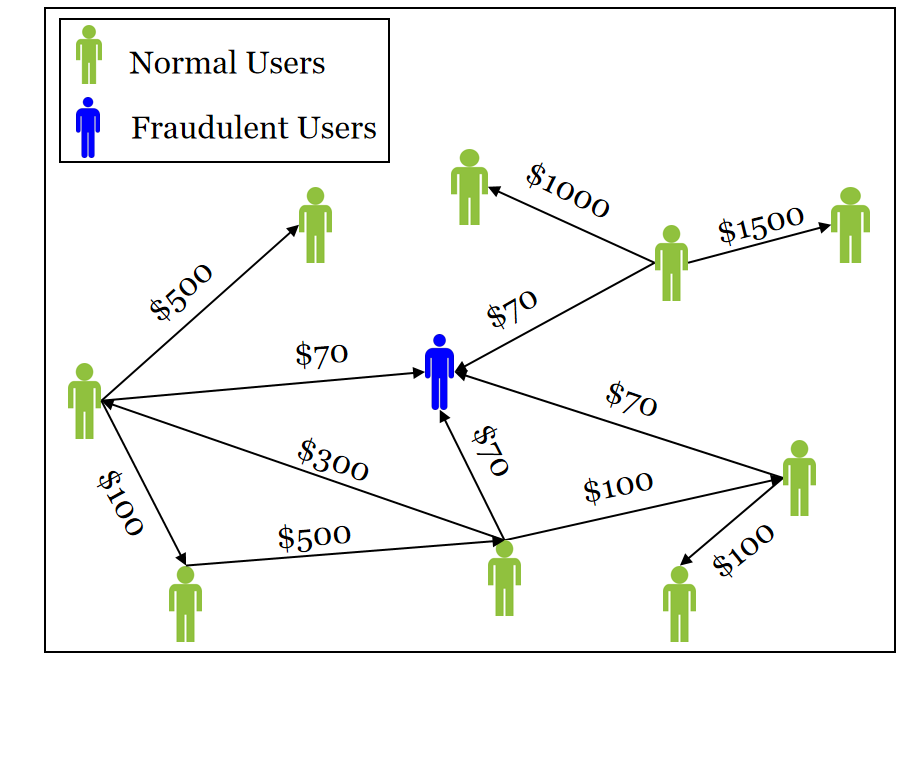}
    }
\hspace{-3.5mm}
\subfigure[Network Layout]{
    \includegraphics[width=4cm, height=3.5cm]{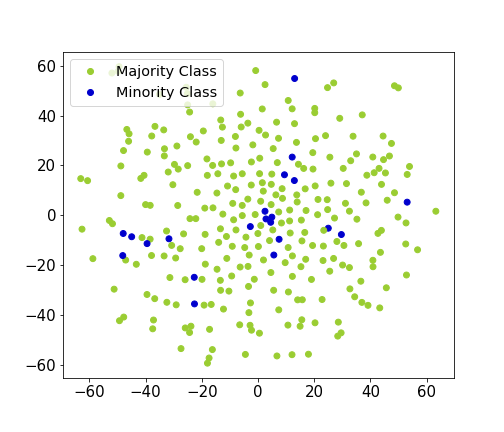}
    }
    
\subfigure[Fraudulent users (over-confident)]{
    \includegraphics[width=4cm, height=3.1cm]{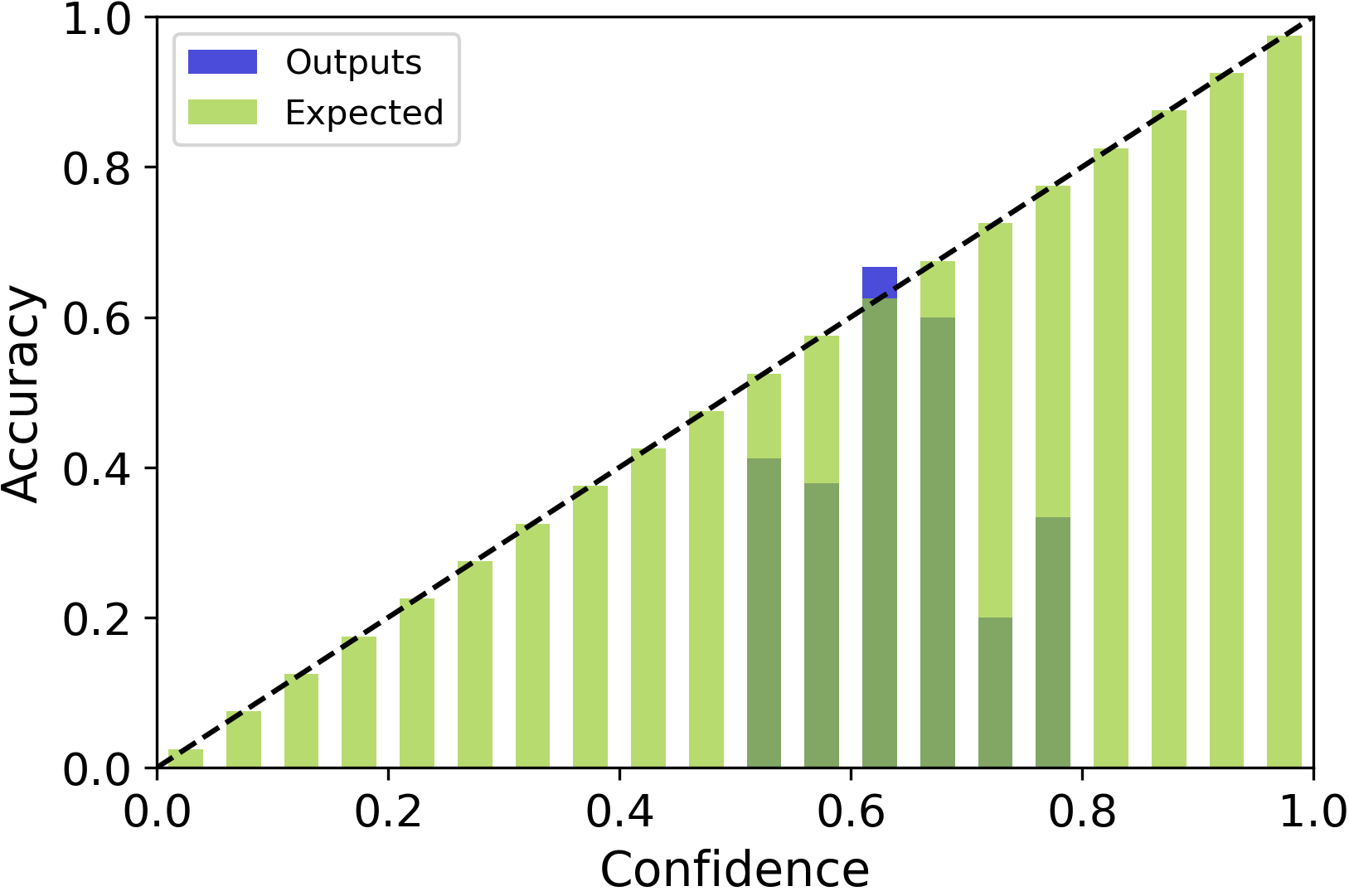}
    }
\hspace{-3.5mm}
\subfigure[Normal users (under-confident)]{
    \includegraphics[width=4cm, height=3.1cm]{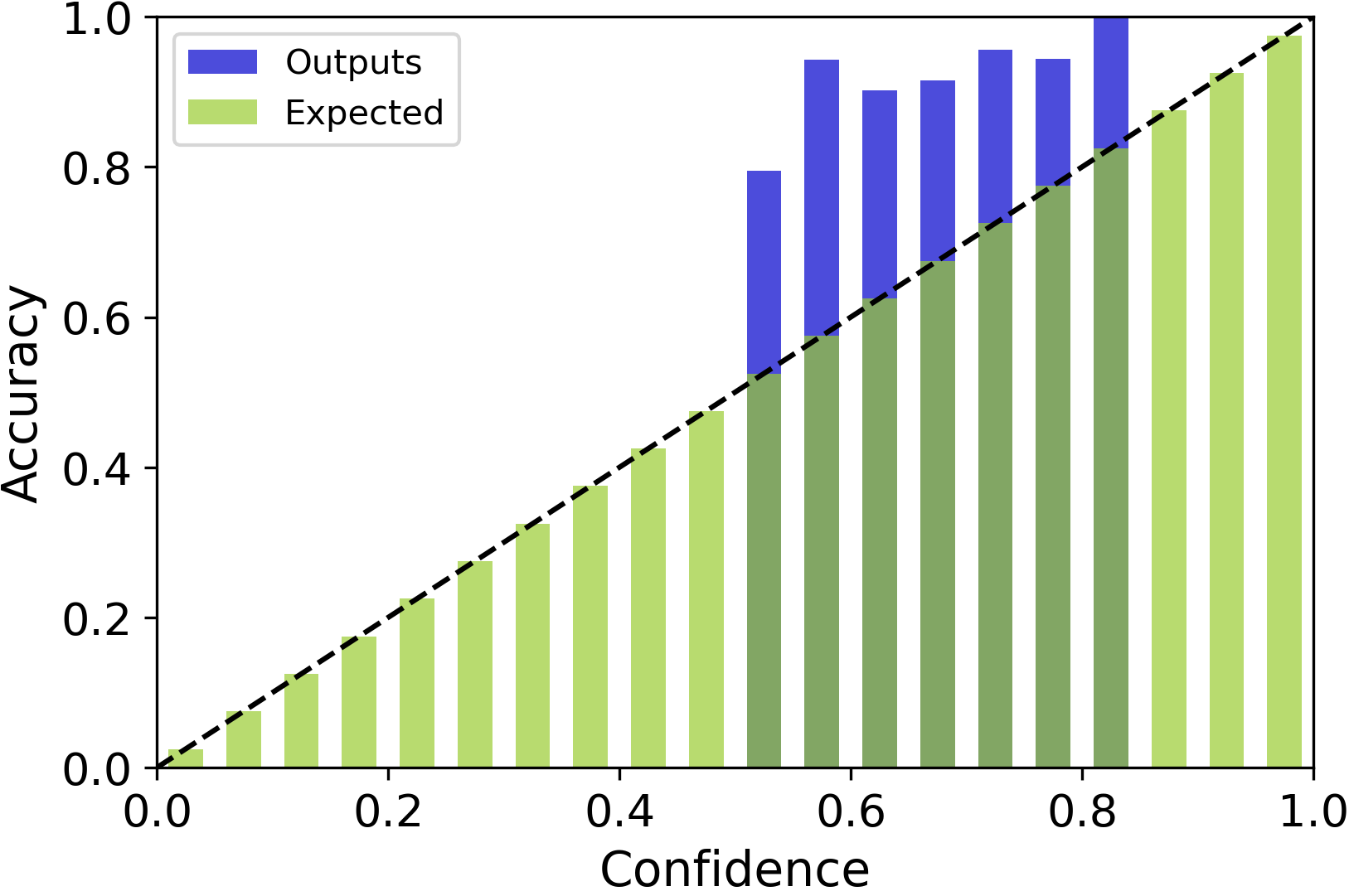}
    }
\caption{An illustrative example of financial fraud detection. (a) is a miniature of a financial transaction network, where 
the blue people are fraudulent users, and the green people are normal users. (b) shows the network layout of a financial transaction network\cite{assefa2020generating}, where fraudulent users colored in blue and normal users colored in green are overlapped together. (c) and (d) are reliability diagrams for the task of predicting fraudulent users and normal users, correspondingly.}
\label{fig: illustration of financial fraud detection} 
\end{figure}

Recent developments in confidence calibration~\cite{niculescu2005predicting, kumar2018trainable, widmann2019calibration, nixon2019measuring, zhang2020mix, minderer2021revisiting} provide a useful tool to make machine learning models uncertainty-aware and thus produce reliable predictions. However, confidence calibration in rare category characterization is non-trivial due to the unique challenges of rare categories. First, (\textbf{C1. Uncertainty}) rare categories are often non-separable minority classes~\cite{zhou2015rare}, such as the fraud patterns shown in Figure~\ref{fig: illustration of financial fraud detection} (b). That is to say, overlapping support regions between classes introduce a large uncertainty in prediction. This is further confirmed by the reliability diagrams in Figure ~\ref{fig: illustration of financial fraud detection} (c) and (d). The x-axis is the confidence interval [0, 1] grouped into 20 equal-size bins, the y-axis is the average accuracy of each bin, and samples are assigned to the corresponding bin 
according to their confidence. A model is reliable and trustworthy if the average confidence of each bin closely matches its accuracy, any deviation from the diagonal represents a miscalibration (or unreliable). However, from Figures~\ref{fig: illustration of financial fraud detection} (c) and (d), we can see that fraudulent users are over-confident, while normal users are under-confident - neither of the predictions is trustworthy. Second (\textbf{C2. Rarity}), it is often the case that the observed data exhibits highly-skewed class-membership distributions, \ie, rare categories are scarce and overwhelmed by other classes. While conventional confidence calibration measures are distribution-based (\eg, expected calibration error (ECE)), they do not measure the calibration error of predictions for individual samples. It could be problematic for calibrating the predictions of rare categories, as the distribution-based confidence calibration measures (\eg, ECE) can not produce accurate estimations when very few or even no rare category samples are observed in each confidence interval.

In this paper, we start with the investigation of the miscalibration of existing rare category characterization methods, finding that (1) existing methods tend to be over-confident in predicting rare categories, and (2) distribution-wise calibration models do not accurately measure miscalibration in the presence of rare categories. Motivated by these observations, we propose for the first time an individual calibration framework named \name\ for graph-structured data, which enables the confidence calibration of rare category analysis at an instance-level granularity. In particular, to address C1, we develop a node-level uncertainty quantification algorithm to model the overlapping support regions with high uncertainty; to address C2, we generalize the distribution-wise confidence calibration metric ECE to the instance level by proposing a novel measurement named Expected Individual Calibration Error (EICE). We conduct extensive experiments on real-world datasets, demonstrating the effectiveness of EICE in rare category characterization in terms of both confidence calibration and prediction performance.
We summarize our contributions as follows:

\begin{itemize}
    \item {\bfseries Problem:} We identify for the first time the calibration problem for rare categories and study the unique challenges motivated by practical applications.  
    \item {\bfseries Model:} We propose an end-to-end framework, called \name, that jointly learns the characterizations of rare categories and calibrates the confidence.
    \item {\bfseries Evaluation:} We systematically evaluate the performance of our method in two settings: 1) confidence calibration and 2) rare category characterization. Extensive results demonstrate the superior performance of our method.
\end{itemize}
The rest of this paper is organized as follows. We provide the problem definition in Section 2, followed by the proposed framework in Section 3. Section 4 discusses the experimental setup and results, followed by a literature review in Section 5. Finally, we conclude the paper in Section 6.

\section{Problem Definition}
In this section, we introduce the preliminaries of our problem setting. Then, we give the formal definition of calibrated on rare category characterization.
We use upper case calligraphic font letters to denote sets (e.g., $\mathcal{G}$), bold upper case letters to denote matrices (e.g., $\bm{A}$), bold lower case letters to denote vectors (e.g., $\bm{x}$), and regular lower case letters to denote scalars (e.g., $\alpha$). 
We use superscript $^T$ for matrix transpose and superscript $^{-1}$ for matrix inversion (e.g., $\bm{A}^T$ and $\bm{A}^{-1}$ are the transpose and inversion of $\bm{A}$ respectively).

\textbf{Rare Category Characterization (RCC).}
Given an undirected attributed graph $\mathcal{G} = (\mathcal{V}, \mathcal{E}, \bm{X})$ with the adjacency matrix $\bm{A} \in \mathbb{R}^{N\times N}$ and the node attribute matrix $\bm{X}=[\bm{x}_1,\cdots,\bm{x}_N]^T$, where $\mathcal{V}$ is the set of nodes, $\mathcal{E}$ is the set of edges, and $N=|\mathcal{V}|$ is the number of nodes.
Moreover, let $\mathcal{K} = \{v_1, \cdots, v_n\} \in \mathcal{V}$ denote $n$ labeled nodes, where we assume that (1) $n < N$ and (2) there is at least one node from each class. 
In the setting of rare category characterization\cite{he2010rare}, the goal is to learn a prediction function $f(\cdot)$ that well characterizes the support regions of the minority classes and outputs a list of predicted rare category samples with high accuracy.

\textbf{Confidence Calibration.} In general, confidence calibration aims to calibrate the confidence (or prediction probability) of the model to let the calibrated confidence directly reflects the probability of the prediction being correct. Mathematically, given a random variable $X$ representing the input feature, $Y$ representing the label, and a model $f(\cdot)$ that outputs the predicted label $\hat{Y}$ and its associated confidence $\hat{P}$. The model $f(\cdot)$ is perfectly calibrated if:
\begin{equation}
\label{eqn:subproblems}
\begin{split}
\mathbb{P}(\hat{Y}=Y|\hat{P}=p) = p, \forall p \in [0, 1]
\end{split}
\end{equation}
 where $\mathbb{P}(\cdot,\cdot)$ represents the joint distribution of the confidence $P$ and label $Y$. 
 The left-hand side denotes the true data distribution's probability of getting a correct label for every sample, the right-hand side denotes the probability. 
 Any difference between them is known as the calibration error.

Many efforts have been made to measure the calibration error. Expected calibration error ~\cite{naeini2015obtaining} is a commonly used metric that approximates the calibration error in expectation. Specifically, it divides prediction probability into a fixed number of bins with equal intervals. The calibration error is the weighted average of the difference between every bin's accuracy and confidence. Formally speaking, the ECE is defined as follows:
\begin{equation}
\label{eqn:subproblems}
\begin{split}
ECE=\sum_{m=1}^{M}\frac{|B_m|}{N}|acc(B_m)-conf(B_m)|,
\end{split}
\end{equation}
where $N$ is the number of samples, $M$ is the number of bins, $|B_m|$ is the number of predictions in the bin $B_m$, $acc(B_m)$ is the accuracy of bin $B_m$, and $conf(B_m)$ is the average confidence of samples in bin $B_m$.
However, the highly-skewed distribution of prediction probability makes only a few bins contribute to ECE~\cite{guo2017calibration}, and the good performance of a bin may be due to the cancellation effect resulting from the overlap of over-confident and under-confident among many predictions ~\cite{nixon2019measuring}. Thus, adaptive calibration error (ACE) is proposed to address these issues. It adaptively chooses the bin intervals to ensure each bin contains the same number of predictions. More precisely,
\begin{equation}
\label{eqn:subproblems}
\begin{split}
ACE=\frac{1}{CM}\sum_{c=1}^{C}\sum_{m=1}^{M}\frac{|B_m|}{N}|acc(B_m, c)-conf(B_m, c)|,
\end{split}
\end{equation}
where $N$ is the number of samples, $C$ is the number of class labels, $M$ is the number of bins, and $|B_m|$ is the number of predictions in the bin $B_m$. The index bin of each prediction is determined by the $\lfloor N/M \rfloor$.

Reliability plays a pivotal role in the development of modern deep learning, especially for high-stakes domains. However, reliable rare category characterization has received little attention. Back to the example of financial fraud detection in Figure~\ref{fig: illustration of financial fraud detection}, the model aims to detect fraud patterns and find fraudulent users from huge amounts of transactions. If the model is over-confident, \ie, the confidence is larger than its accuracy, it may misclassify normal users as fraudulent users, giving users a bad experience and damaging the financial institution's reputation. Otherwise, if the model is under-confident, \ie, the confidence is lower than its accuracy, it may miss the fraudulent users and fail to prevent illicit activities. Thus, the method not only needs to be accurate but also needs to indicate when it is likely to be incorrect. Here, we give the formal definition of calibrated on rare category characterization in Problem 1.

{\setlength{\parindent}{0pt}
\begin{problem}
 \textbf{Calibration on rare category characterization} \\
    \textbf{Input:} (i) An undirected attributed graph $\mathcal{G} = (\mathcal{V}, \mathcal{E})$ with the adjacency matrix $\bm{A}$, the node attribute matrix $\bm{X}$, and the label $\bm{Y}$, (ii) a rare category characterization model $f(\cdot)$, (iii) the outputs of $f(\cdot)$ including the prediction labels $\hat{Y}$ and the corresponding prediction confidence $P$. \\
   \textbf{Output:} The calibrated confidence $\hat{P}$ of the classifier $f(\cdot)$.
   \end{problem}
}

\begin{figure*}[h]
\small
\captionsetup{font=footnotesize}
\centering
\subfigure[All]{
    \includegraphics[width=5.6cm, height=4.0cm]{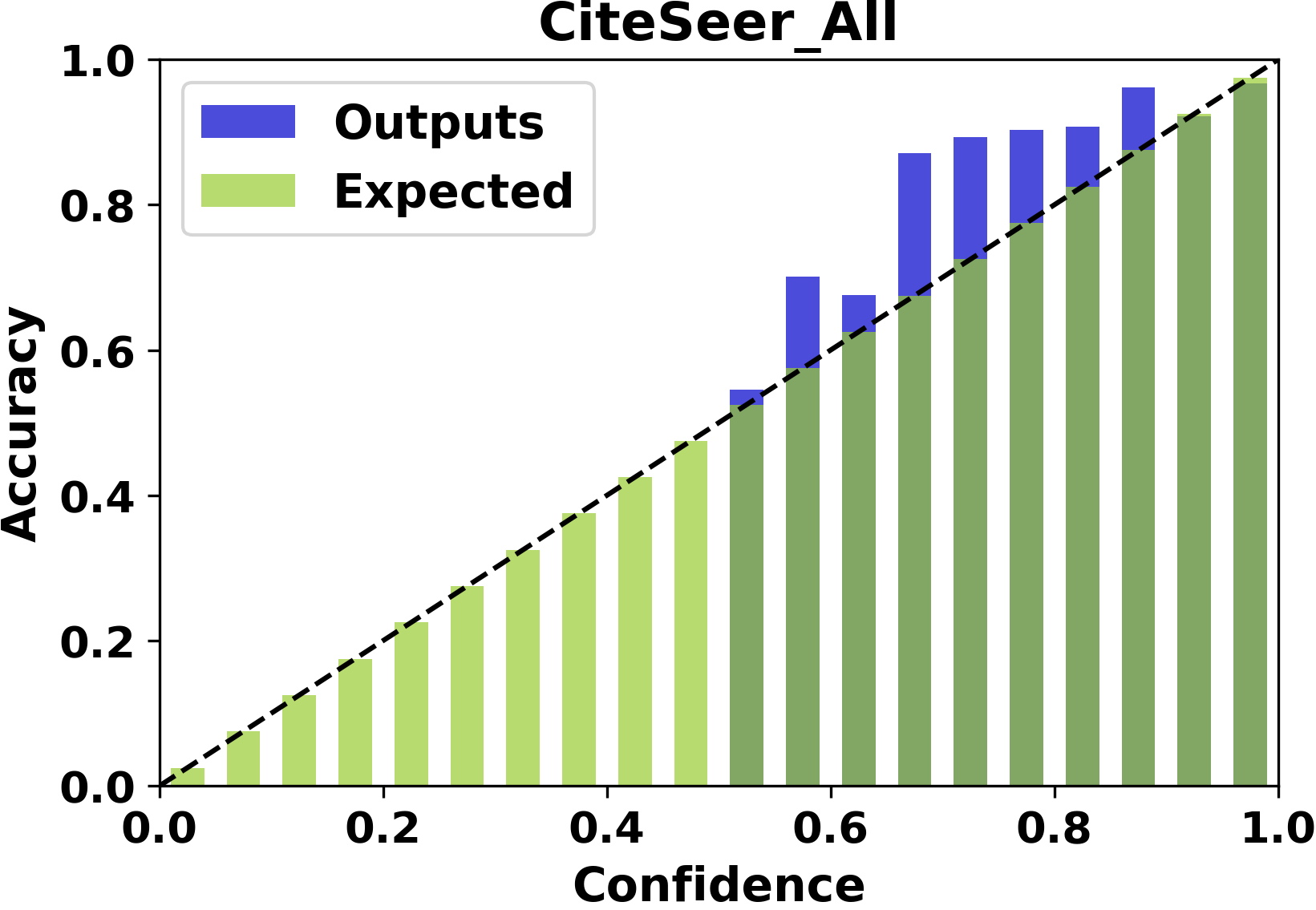}
    }
\subfigure[Majority]{
    \includegraphics[width=5.6cm, height=4.0cm]{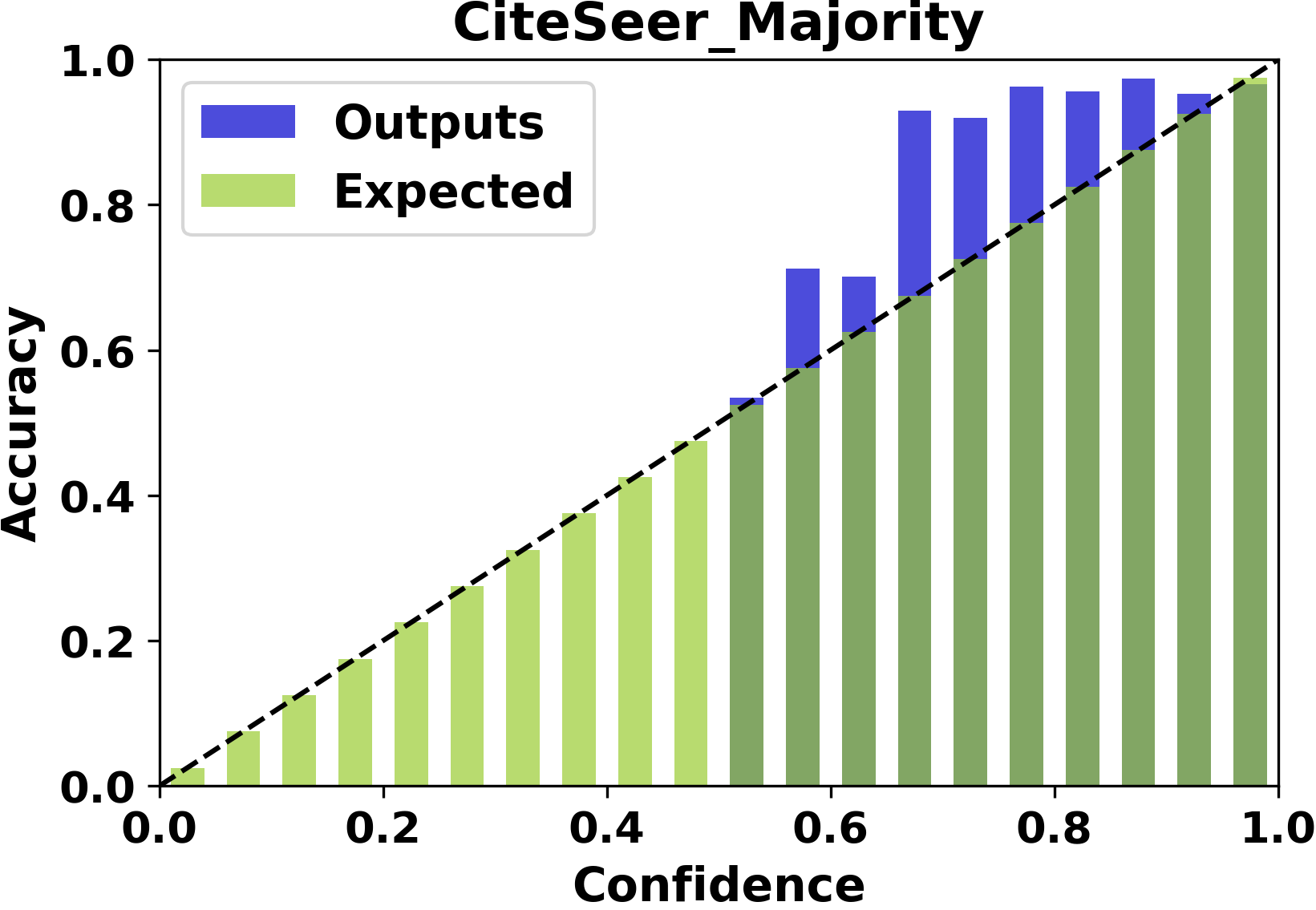}
    }
\subfigure[Minority]{
    \includegraphics[width=5.6cm, height=4.0cm]{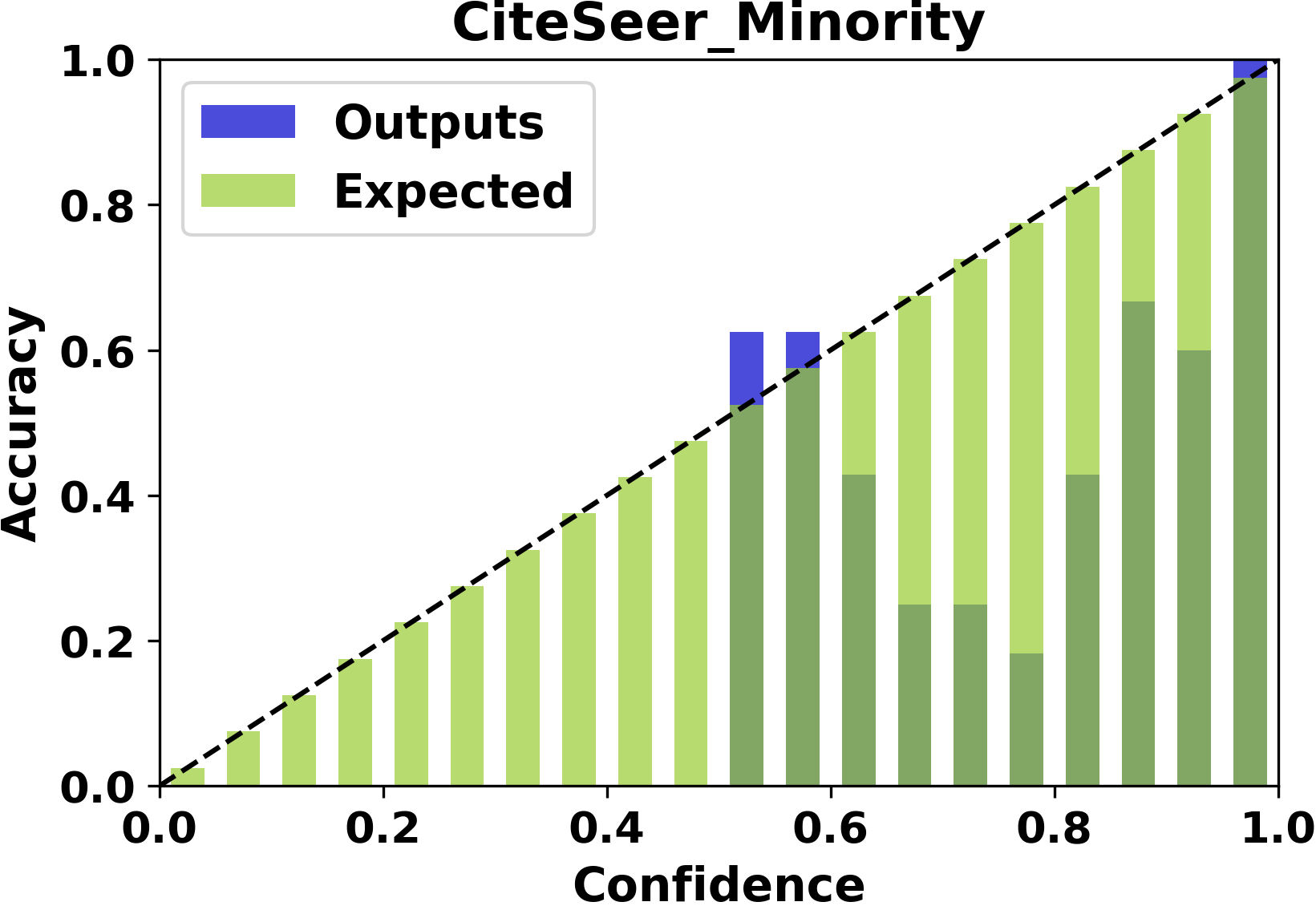}
    }
\subfigure[All]{
    \includegraphics[width=5.6cm, height=4.0cm]{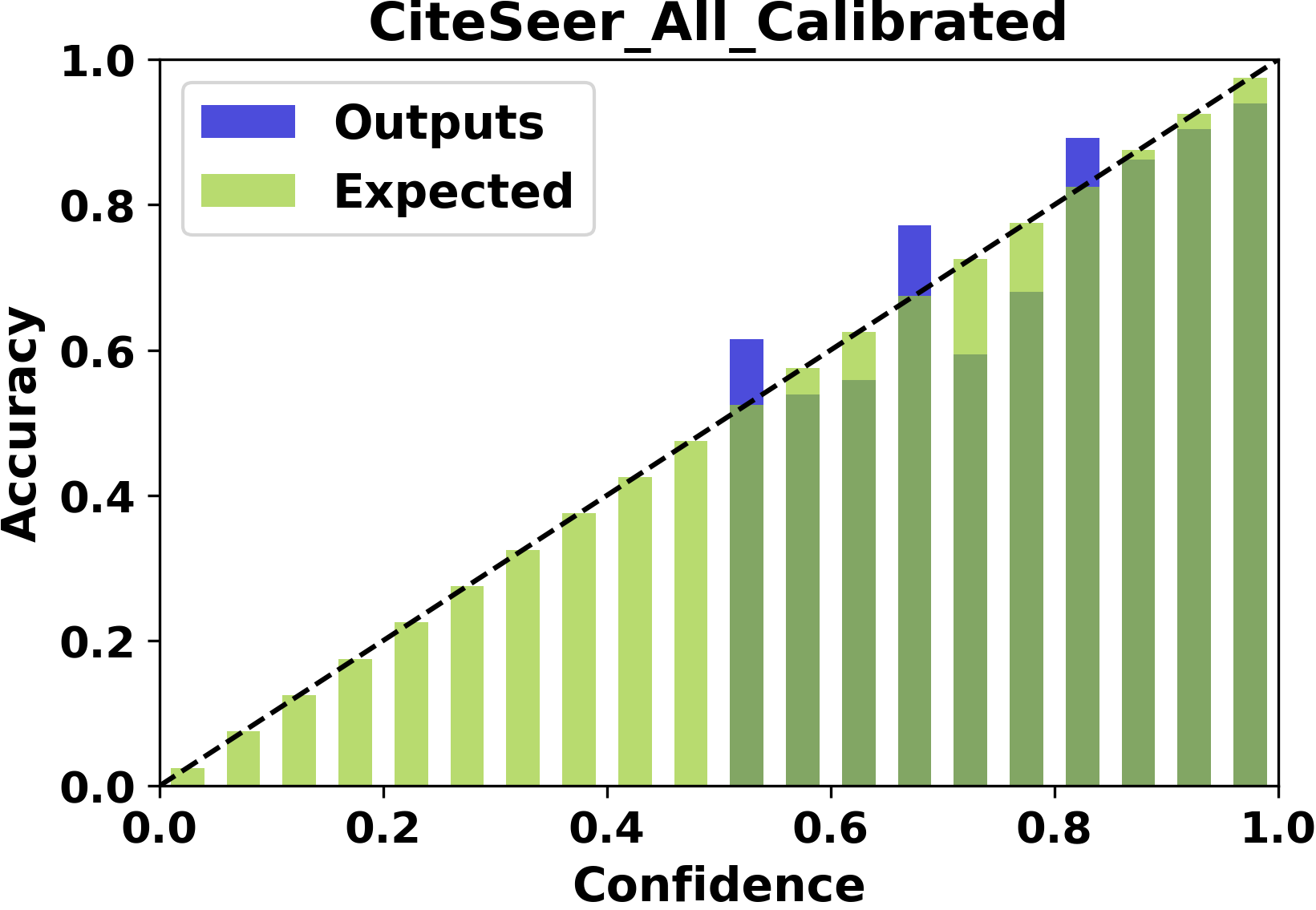}
    }
\subfigure[Majority]{
    \includegraphics[width=5.6cm, height=4.0cm]{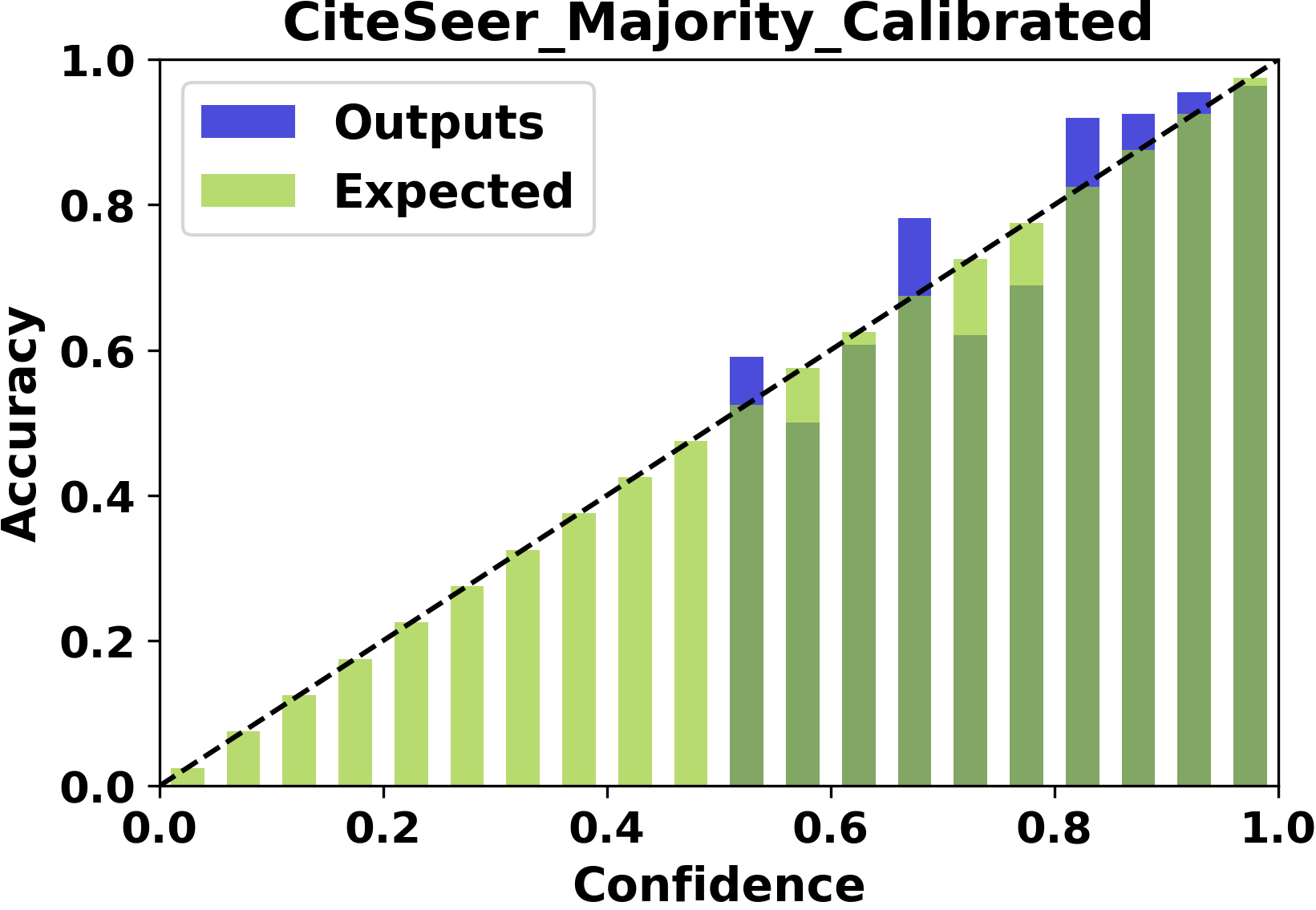}
    }
\subfigure[Minority]{
    \includegraphics[width=5.6cm, height=4.0cm]{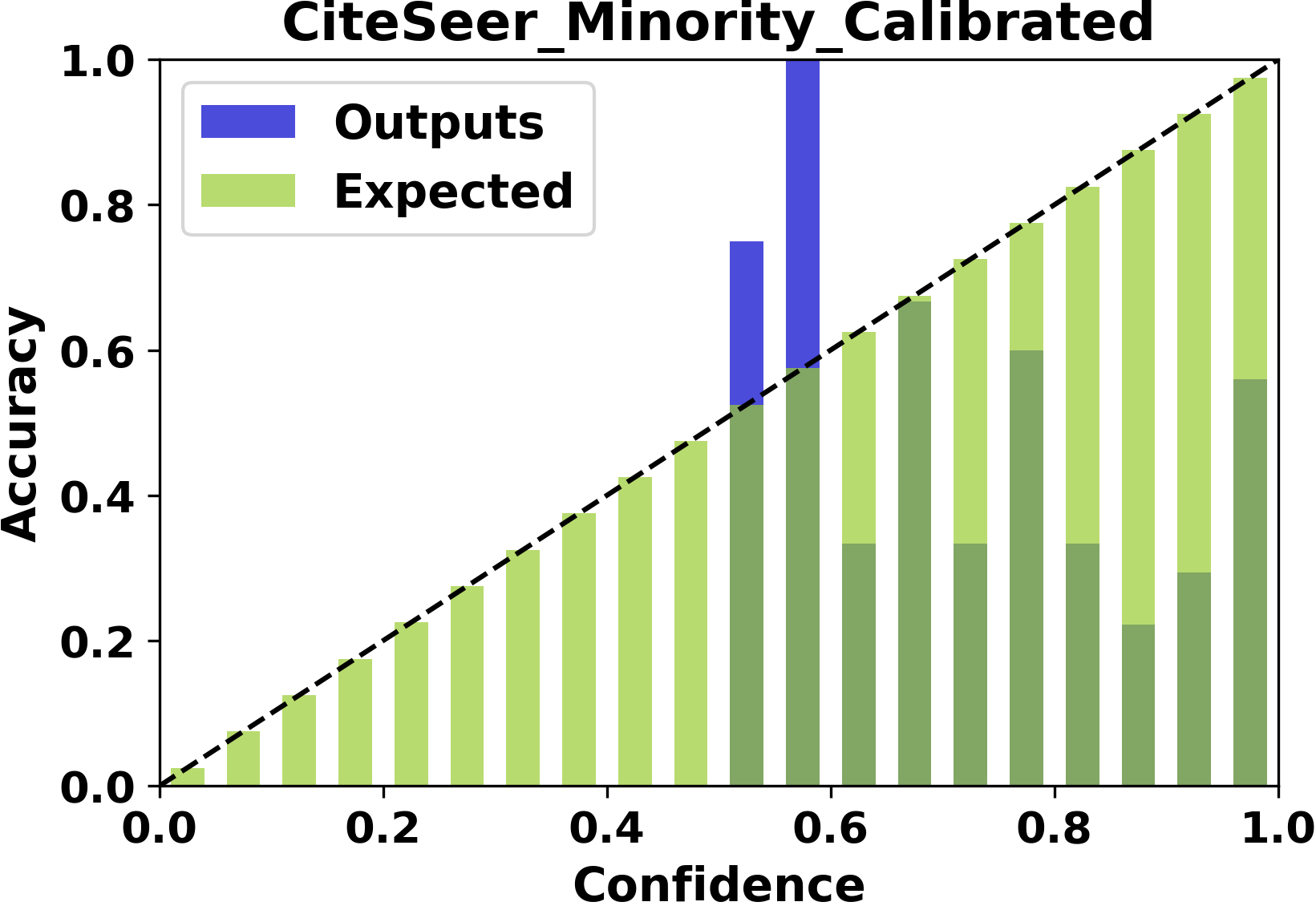}
    }
\caption{Reliability diagrams for GCN on CiteSeer w/o calibration. The confidence of a well-calibrated model should closely match its accuracy, namely, aligned with the diagonal. Below the diagonal represents over-confident, and above the diagonal represents under-confident.}
\label{fig: histogram of CiteSeer} 
\end{figure*}

\section{Calibration on Rare Category}
In this section, we introduce our proposed framework \name\ for confidence calibration on rare category characterization.
We start by investigating the risk of miscalibration in predicting rare categories, which reveals that the state-of-the-art methods are primarily over-confident in predicting rare categories.
Then, we introduce a novel individual calibration metric named EICE that is designed to alleviate the miscalibration in rare category analysis. 
Finally, we present the theoretical analysis of EICE and provide an end-to-end optimization algorithm \name\ for rare category calibration.

\subsection{Over-Confident Tendency in Predicting Rare Categories} 
Recall the reliability diagrams in Figure~\ref{fig: illustration of financial fraud detection}, which reveals that the existing RCC methods are primarily over-confident in predicting minority classes while under-confident in predicting majority classes. 
This observation naturally gives rise to two questions: (1) Does this observation universally true? and (2) Can we apply existing calibration tools to alleviate the miscalibration in rare category characterization? 

To answer these questions and have a better understanding of Problem 1, we further conduct a case study on a collaboration network~\cite{sen2008collective, DBLP:conf/kdd/ZhouZ0H20}, which is composed of one majority class and one minority class. Specifically, we consider the cost-sensitive GCN (CGCN) ~\cite{pan2015cogboost} as the baseline method \footnote{For binary classification problems with positive class ($+$) and negative class ($-$), cost-sensitive learning is to learn a model with minimum misclassification costs: $C(-,+) \times \#FN + C(+,-)\times \#FP$, where $C(i,j)$ is the cost of predicting a sample belonging to class $i$ when in fact it belongs to class $j$, $\#FN$, and $\#FP$ are the numbers of false negative and false positive samples respectively.}, which is widely used in rare category characterization. 
In Figure~\ref{fig: histogram of CiteSeer}, we present the reliability diagram, where the x-axis represents the confidence intervals with equal size, and the y-axis represents the average accuracy in each bin. 
In particular, the three figures in the first row of Figure~\ref{fig: histogram of CiteSeer} show the results of CGCN without calibration, while the three figures in the second row of Figure~\ref{fig: histogram of CiteSeer} show the results of CGCN with calibration (\ie, temperature scaling~\cite{guo2017calibration}).
The blue bars show the empirical accuracy of CGCN, while the green bars correspond to the expected accuracy based on confidence. Intuitively, if the empirical accuracy of CGCN is equal to the expected accuracy in each bin, then we can claim that the model is perfectly calibrated. 
However, in Figure~\ref{fig: histogram of CiteSeer}, we observe that 
(1) CGCN is over-confident in predicting minority classes while under-confident in predicting majority classes, which is consistent with our observations in Figure~\ref{fig: illustration of financial fraud detection}.
(2) By comparing Figure~\ref{fig: histogram of CiteSeer} (b) and (e), temperature scaling achieves good performance in calibrating CGCN in predicting the majority class.
(3) By comparing Figure~\ref{fig: histogram of CiteSeer} (c) and (f), temperature scaling not only fails to calibrate CGCN in predicting the minority class but even amplifies the miscalibration in Figure~\ref{fig: histogram of CiteSeer} (f). 

We believe two factors result in the miscalibration phenomenon for RCC: characterizations of rare categories and calibration metrics.
First, rare categories are naturally scarce. The distribution of rare categories is highly skewed, and they are often overwhelmed by the majority class, so little attention has been paid to the rare categories during the process of calibration. Additionally, rare categories are often non-separable from the majority class. The overlapping support regions between rare categories and majority classes introduce uncertainty, which inevitably reduces the reliability of the model and increases miscalibration.
Second, the widely-used metric for calibration, ECE, suffers from several issues ~\cite{kumar2019verified, nixon2019measuring}. 
On the one hand, the highly-skewed distribution of predictions. One desired property of a probabilistic model is sharpness, which means the model should always predict with high confidence. This property leads to the result that most predictions are assigned to the right bins, while few predictions are assigned to the left bins. For the bins on the left, the number of predictions is very small and their accuracy is inaccurate and prone to change, bringing inaccurate calibration errors. For the bins on the right, the number of predictions for each bin is so large that rare categories are ignored when calculating the average accuracy and confidence.
Additionally, there exists an overlap of over-confident and under-confident between lots of predictions in the same bin. Thus, the calibration error may be reduced due to the cancellation effect, not the true improvement. 
On the other hand, the selection of the number of bins. More bins will alleviate the cancellation effect but inevitably introduce more noise, while fewer bins approximate accuracy more precisely but bring more overlap between predictions. The number of bins directly determines the number of predictions in each bin and therefore affects their calibration errors.

\subsection{Expected Individual Calibration Error}
Following the discussion in subsection 3.1, the core issue of ECE is that this metric is distribution-based, selecting the number of bins will inevitably make mistakes. To address this challenge, we propose a novel measurement: Expected Individual Calibration Error (EICE), which approximates calibration error to the instance level and does not need to choose the number of bins. Ideally, we expect every sample to be perfectly calibrated and the final result of EICE is 0. The formal definition of EICE is given as follows:
\begin{equation}
\label{eqn: expected ice}
\begin{split}
\mathbb{E}_{\hat{p}}[|\mathbb{P}(\hat{y}=y|\hat{p}=p) - p|].
\end{split}
\end{equation}
where $\hat{y}$ is the predicted label of the sample and $\hat{p}$ is its associate confidence. Different from ECE which approximates expectations by grouping predictions into bins, EICE measures calibration error at the level of individual samples.  
Since the accuracy for each sample is either 0 (incorrect) or 1 (correct), which cannot indicate the true probability of getting a correct prediction precisely. 
Recent advances in uncertainty quantification demonstrate that uncertainty quantification could reflect how uncertain the predictions are and discriminate between high and low-confidence predictions~\cite{alaa2020discriminative}, i.e., jackknife quantifies uncertainty in terms of the average prediction error~\cite{barber2021predictive}.
Thus, we propose to utilize the sample's uncertainty to represent the individual prediction performance of $f(\cdot)$.  

Inspired by~\cite{kang2022jurygcn}, we leverage the jackknife uncertainty to quantify the uncertainty of samples. Let $f(v; \theta)$ denote the model, where $v$ is a node and $v \in \mathcal{V}$, $\theta \in \Theta$ are the model parameters, and $\Theta$ is the parameter space. The parameters $\hat{\theta}$ are trained based on solving the optimization problem:
\begin{equation}
\label{eqn:optimization}
\begin{split}
\hat{\theta} = \arg \min_{\theta \in \Theta} \mathcal{L}(\mathcal{V}_{train};\theta)
\end{split}
\end{equation}
where $\mathcal{V}_{train} \subseteq \mathcal{V}$ is the set of training nodes, $\mathcal{L}(\mathcal{V}_{train},\theta)$ is the loss of training set, such as cross-entropy, etc.

The jackknife estimates the node-level confidence interval $\mathcal{C}(v;\theta)$ by the technique of leave-one-out (LOO). More precisely, it leaves out each sample in $\mathcal{V}_{train}$ to re-train the model and evaluates the performance of every re-trained model on the held-out sample $v$. The $\mathcal{C}(v;\hat{\theta})$ is defined as follows:
\begin{equation}
\label{eqn: confidence interval}
\begin{split}
\mathcal{C}_\alpha(v;\hat{\theta}) = [\mathcal{Q}_\alpha(\{f(v; \hat{\theta_{-i}}) - r_i\}), \mathcal{Q}_{1-\alpha}(\{f(v; \hat{\theta_{-i}}) + r_i\})]
\end{split}
\end{equation}
where $r_i = |y_i - f(i; \hat{\theta_{-i}})|$ is the error residual of node $i \in \mathcal{V}_{train}$, $\hat{\theta_{-i}}$ are the parameters of the model re-trained on the dataset $\mathcal{V}_{train} \backslash \{i\}$ (remove node $i$ from training set), $\mathcal{Q_\alpha}(\mathcal{R})$ and $\mathcal{Q}_{1-\alpha}(\mathcal{R})$ are the $\alpha$ and $1-\alpha$ quantile of the set $\mathcal{R}$,
and $\mathcal{Q}_\alpha(\mathcal{R}) = \mathcal{Q}_{1-\alpha}(-\mathcal{R})$.

However, the re-training procedure is time-consuming and computationally expensive, especially for large-scale datasets. Through the influence functions~\cite{koh2017understanding} (essentially Taylor expansion over the model parameters), the parameters $\hat{\theta_{-i}}$ can be obtained based on the trained model $f(v; \hat{\theta)}$ without re-training the model. The time complexity is linear in $O(Np)$, where $p$ is the number of parameters of the model, and $N$ is the size of the training set~\cite{alaa2020discriminative}.

If node $i$ was upweighted by a small perturbation $\epsilon$, the new parameters $\hat{\theta_{i,\epsilon}}$ are as follows:
\begin{equation}
\label{eqn:influence_function}
\begin{split}
\hat{\theta_{i,\epsilon}} =& \arg \min_{\theta \in \Theta} \mathcal{L}(\mathcal{V}_{train};\theta) + \epsilon \mathcal{L}(i;\theta) \\
& \approx \hat{\theta_{i}} + \epsilon (\mathcal{H}^{-1}_{\hat{\theta}}\nabla_{\theta}L(i;\hat{\theta}))
\end{split}
\end{equation}
where $\mathcal{H}_{\hat{\theta}}=\frac{1}{|\mathcal{V}_{train}|}\nabla^2_{\theta}\sum_{i}L(i;\hat{\theta})$ is the Hessian matrix. For simplicity, we derive an approximate expression for the first-order influence function.

After applying influence functions, the parameters $\hat{\theta_{-i}}$ can be approximately obtained by setting $\epsilon=-\frac{1}{|\mathcal{V}_{train}|}$, since removing a node is equivalent to upweight it by $-\frac{1}{|\mathcal{V}_{train}|}$.

With Eq.\ref{eqn: confidence interval} and Eq.\ref{eqn:influence_function}, we get the lower bound $\mathcal{C}_{\alpha}^-(v;\hat{\theta})$ and upper bound $\mathcal{C}_{\alpha}^+(v;\hat{\theta})$ of the confidence interval with respect to node $v$ without re-training the model. The definition of the uncertainty of node $v$ is defined as follows:
\begin{equation}
\label{eqn:uncertainty}
\begin{split}
\mathcal{C}_{\alpha}^-(v;\hat{\theta}) =& \mathcal{Q}_\alpha(\{f(v; \hat{\theta_{-i}}) - r_i\}) \\
\mathcal{C}_{\alpha}^+(v;\hat{\theta}) =& \mathcal{Q}_{1-\alpha}(\{f(v; \hat{\theta_{-i}}) + r_i\}) \\
uncer(v;\hat{\theta})=& \frac{\mathcal{C}_{\alpha}^-(v;\hat{\theta}) + \mathcal{C}_{\alpha}^+(v;\hat{\theta})}{2}
\end{split}
\end{equation}

With the uncertainty and confidence of a node, we approximate individual calibration error by the difference between its uncertainty and confidence. More precisely, 
\begin{equation}
\label{eqn:ice}
\begin{split}
ICE(v) = |uncer(v;\hat{\theta}) - conf(v;\hat{\theta})|
\end{split}
\end{equation}

Formally, given an L-layer GCN model $f(\cdot)$ with parameters $\hat{\theta}$ and the uncertainty of nodes, the EICE is defined as follows:
\begin{equation}
\label{eqn:ece}
\begin{split}
EICE =\sum_{v \in \mathcal{V}}\frac{1}{N} ICE (v) = \sum_{v \in \mathcal{V}}\frac{1}{N} |uncer(v;\hat{\theta}) - conf(v;\hat{\theta})|
\end{split}
\end{equation}
where $uncer(v;\hat{\theta})$ is the uncertainty of node $v$, and $conf(v;\hat{\theta})$ is the confidence of node $v$. 

The detailed description is presented in Algorithm ~\ref{Alg1}. In particular, the inputs include an undirected attributed graph $\mathcal{G} = (\mathcal{V}, \mathcal{E})$ with the training set $\mathcal{V}_{train}$ and validation set $\mathcal{V}_{val}$, a classifier model $f(\cdot)$ with parameters $\hat{\theta}$, loss function $\mathcal{L}$, and the coverage parameter $\alpha$. 
For Step 1 to Step 5, we compute the new model parameters $\hat{\theta}_{-i}$ after removing node $i$ from the training set, and the error residual $r_i$ for every node $i \in \mathcal{V}_{train}$. 
For Step 6 to Step 11, we first compute the lower bound and upper bound for every node $v \in \mathcal{V}_{val}$ according to $\hat{\theta}_{-i}$ and $r_i$. Then, we compute the uncertainty and confidence of every node $v$. With $uncer(v, \hat{\theta})$ and $conf(v, \hat{\theta})$, we get the $ICE(v)$ being the difference between them.
Finally, for step 12, we output the EICE by taking the average ICE from every node in the validation set.

\renewcommand{\algorithmicrequire}{\textbf{Input:}}
\renewcommand{\algorithmicensure}{\textbf{Output:}}
\begin{algorithm}[t]
\caption{Expected individual calibration error (EICE)}
\label{Alg1}
\begin{algorithmic}[1]
\Require \item[]
    An undirected attributed graph $\mathcal{G} = (\mathcal{V}, \mathcal{E})$ with training set $\mathcal{V}_{train}$ and validation set $\mathcal{V}_{val}$, a classifier model $f(\cdot)$ with parameters $\hat{\theta}$ and cross-entropy loss $\mathcal{L}$, coverage parameter $\alpha$.
\Ensure \item[]
    The EICE.

\For{every $i \in \mathcal{V}_{train}$}
        \State
        \parbox[t]{\dimexpr\linewidth-\algorithmicindent}{
        Compute the Hessian matrix $\mathcal{H}_{\hat{\theta}} \leftarrow \frac{1}{|\mathcal{V}_{train}|}\nabla^2_{\theta}\sum_{i}L(i;\hat{\theta})$.
        }
        \State
        \parbox[t]{\dimexpr\linewidth-\algorithmicindent}{
        Compute new model parameters \\
        $\hat{\theta_{-i}} \leftarrow \hat{\theta} -\frac{1}{|\mathcal{V}_{train}|} \mathcal{H}^{-1}_{\hat{\theta}}\nabla_{\theta}L(i;\hat{\theta})$.
        }
        \State 
        \parbox[t]{\dimexpr\linewidth-\algorithmicindent}{
        Compute the error residual $r_i \leftarrow |y_i - f(i; \hat{\theta_{-i}})|$  .}
\EndFor

\For{every $v \in \mathcal{V}_{val}$}
    \State
    \parbox[t]{\dimexpr\linewidth-\algorithmicindent}{
    Compute the confidence interval of node $v$ with lower bound $\mathcal{C}_{\alpha}^-(v;\hat{\theta}) \leftarrow \mathcal{Q}_\alpha(\{f(v; \hat{\theta_{-i}}) - r_i\})$ and upper bound \linebreak $\mathcal{C}_{\alpha}^+(v;\hat{\theta}) \leftarrow \mathcal{Q}_{1-\alpha}(\{f(v; \hat{\theta_{-i}}) + r_i\})$. 
    }
    \State
    \parbox[t]{\dimexpr\linewidth-\algorithmicindent}{
   Compute the uncertainty of node $v$, \linebreak $uncer(v;\hat{\theta}) \leftarrow (\mathcal{C}_{\alpha}^-(v;\hat{\theta}) + \mathcal{C}_{\alpha}^+(v;\hat{\theta}))/2$.
    }
    \State 
    \parbox[t]{\dimexpr\linewidth-\algorithmicindent}{
    Compute the confidence $conf(v;\hat{\theta})$ of the node $v$ for the given classifier model $f(\cdot)$ with parameters $\hat{\theta}$.
    }
    \State 
    \parbox[t]{\dimexpr\linewidth-\algorithmicindent}{
    Compute the individual calibration error of the node $v$ with Equation~\ref{eqn:ice}, $ICE(v) \leftarrow |uncer(v;\hat{\theta}) - conf(v;\hat{\theta})|$.
    }
\EndFor

\State \Return $EICE \leftarrow \sum_{v \in \mathcal{V}_{val}}\frac{1}{|\mathcal{V}_{val}|}ICE(v)$
\end{algorithmic}
\end{algorithm}

\subsection{Theoretical Analysis}
In this section, we prove the theoretical properties of the metrics we proposed: ICE and EICE. 
First, we show that ICE is a faithful measure of the discrepancy between confidence and accuracy, which is zero if and only if the model $f(\cdot)$ is fully calibrated. Next, we compare EICE and ECE and show that a small EICE implies a better calibration compared to a small ECE.

\begin{property}\label{prop-cali}
For the test node $v \in \mathcal{V}$, $ICE(v)= 0$ if and only if the model $f(\cdot)$ is perfectly calibrated for $v$.
\end{property}

\begin{proof}\label{def-cali}
On the one hand, we assume $ICE(v)=|\mathbb{P}(\hat{y}=y|\hat{p}=p) - p| = 0$, where $\hat{p}=p$ represents $conf(v;\theta)=p$, and $\mathbb{P}(\hat{y}=y)$ indicates $acc(v;\theta)$.
Since $|\mathbb{P}(\hat{y}=y|\hat{p}=p) - p|\geq 0$, we have $\mathbb{P}(\hat{y}=y|\hat{p}=p) = p$. Thus, if $conf(v;\theta)=p$, we can get $acc(v;\theta)=p=conf(v;\theta)$, which implies that the model $f(\cdot)$ is perfectly calibrated for $v$. 

On the other hand, we assume the model $f(\cdot)$ is perfectly calibrated for $v$, i.e., $conf(v;\theta) = acc(v;\theta)$. Then, for any $p$, if we have $conf(v;\theta)=p$, we can get $acc(v;\theta)=p$. This indicates that $\mathbb{P}(\hat{y}=y|\hat{p}=p) = p$. Thus, we can get $ICE(v)=|\mathbb{P}(\hat{y}=y|\hat{p}=p) - p| = 0$.
\end{proof}

\begin{remark}
Property~\ref{prop-cali} demonstrates the equivalence between the small ICE and the calibrated individual classification, which implies that ICE is a proper scoring rule. This gives us the opportunity to analyze calibration at the individual level and bring benefits to rare categories.
\end{remark}

\begin{property}\label{prop-strong}
If we have $EICE=0$, we will have $ECE=0$. And conversely, if we have $ECE=0$, we can still have $EICE > 0$.
\end{property}
\begin{proof}
On the one hand, we assume $EICE=0$. According to Property~\ref{prop-cali}, we have $\mathbb{E}_{\hat{p}}[|\mathbb{P}(\hat{y}=y|\hat{p}=p) - p|] = 0$. 
For the ECE, we have 
\begin{equation} 
\label{eqn:expand_ece}
\begin{split}
ECE &=\mathbb{E}_{\hat{p}}[|\mathbb{P}(\hat{Y}=Y|\hat{P}=p) - p|] \\
    &= \mathbb{E}_{\hat{p}}[|\sum_{y \in Y} (\mathbb{P}(\hat{y}=y|\hat{p}=p) - p)|] \\
    &\leq \sum_{y \in Y} \mathbb{E}_{\hat{p}}[| \mathbb{P}(\hat{y}=y|\hat{p}=p) - p|]=0
\end{split}
\end{equation}
Since $ECE\geq 0$, we can get $ECE=0$.

On the other hand, we assume $ECE=0$, i.e., $\mathbb{E}_{\hat{p}}[|\sum_{y \in Y} (\mathbb{P}(\hat{y}=y|\hat{p}=p) - p)|]=0$. Suppose we have two nodes $v_1, v_2 \in \mathcal{V}$ with label $y_1$ and $y_2$, respectively. Consider the case $\mathbb{P}(\hat{y_1}=y_1|\hat{p_1}=p) - p=0.1$ and $\mathbb{P}(\hat{y_2}=y_2|\hat{p_2}=p) - p=-0.1$, the $EICE = 0.2 \neq 0$, but $ECE=0$.
\end{proof}

\begin{remark}
Property~\ref{prop-strong} shows that our EICE is more stringent in terms of model calibration compared to ECE. The previous ECE measures calibration from a global perspective and examines the discrepancy at each confidence level. 
This does not guarantee a well-calibrated classification at the individual level and can make the model prone to ignore rare categories.
\end{remark}

\subsection{Optimization Objective}
In this paper, we consider the task of rare category characterization with an L-layer GCN. We found that the loss of many node-level tasks can be represented as $\mathcal{L}=\frac{1}{|V_{train}|} \sum_{v \in V_{train}} l(v;\theta)$. There are many choices for the loss function $l$, and Cross-Entropy (CE) is the most widely used loss function for classification tasks, which is also utilized in our paper. So the loss can be written as:
\begin{equation} 
\label{eqn:loss_nll}
\begin{split}
\mathcal{L}_{CE} = - \frac{1}{|V_{train}|}\sum_{i=1}^{|V_{train}|}\sum_{c=1}^{C} y_{i}log p_{i}
\end{split}
\end{equation}
where $y_i$ is the one-hot encoding label for node $i$, and $p_i$ is the prediction probability of the model for node $i$.

Due to the different properties of the majority class and minority class of GCNs, we intend to calibrate the model with respect to the EICE. Hence, we design a regularization term as follows:
\begin{equation} 
\small
\label{eqn:loss_iece}
\begin{split}
\mathcal{L}_{EICE} = \frac{1}{|V_{train}|}\sum_{i=1}^{|V_{train}|} ICE(i)
\end{split}
\end{equation}
where $ICE(i)$ is the individual calibration error of node $i$. 

Intuitively, we'd like to improve the performance of the model in terms of calibration and classification. Thus, We design \name\ to jointly learn the characterizations of rare categories and calibrate the confidence. 
Specifically, we combine $\mathcal{L}_{CE}$ and $\mathcal{L}_{EICE}$ together, and the overall learning objective of our method can be rewritten as Eq.~\ref{eqn: loss}. 
\begin{equation} 
\small
\label{eqn: loss}
\begin{split}
\mathcal{L} = (1-\lambda) \mathcal{L}_{CE} + \lambda \mathcal{L}_{EICE}
\end{split}
\end{equation}
where $\lambda$ is the hyper-parameter that controls the trade-off between the performance of calibration and classification.

The detailed description is presented in Algorithm ~\ref{Alg2}. With a classifier model $f(\cdot)$ as the base model, we update its parameters until the overall loss $\mathcal{L}$ converges or reaches the maximum number of epochs.

\begin{algorithm}[t]
\caption{\name\ Algorithm} 
\label{Alg2}
\begin{algorithmic}[1]
\Require \item[]
    Graph $\mathcal{G} = (\mathcal{V}, \mathcal{E})$, a classifier model $f(\cdot)$ with parameters $\hat{\theta}$, the hyper-parameter $\lambda$, a test node $v$.
\Ensure \item[]
    The calibrated prediction probability of test node $v$.

\For{$epoch=1:e$}
        
        \State
        \parbox[t]{\dimexpr\linewidth-\algorithmicindent}{
        Compute the original loss $\mathcal{L}_{CE}$ for model $f(\cdot)$.
        }      
        \State
        \parbox[t]{\dimexpr\linewidth-\algorithmicindent}{
        Compute the regualarizer $\mathcal{L}_{IECE}$ using Algorithm ~\ref{Alg1}.
        }
        \State
        \parbox[t]{\dimexpr\linewidth-\algorithmicindent}{
        Compute the overall learning objective $\mathcal{L}$ based on the combination of $\mathcal{L}_{CE}$ and $\mathcal{L}_{ICE}$ from Equation\ref{eqn: loss}.
        }
        \State 
        \parbox[t]{\dimexpr\linewidth-\algorithmicindent}{
        If the overall loss $\mathcal{L}$ converges, break; otherwise, update the parameters $\hat{\theta}$.}
\EndFor
\State \Return The calibrated prediction probability of test node $v$ by model $f(\cdot)$ with updated parameters $\hat{\theta}$. 
\end{algorithmic}
\end{algorithm}

\section{Experiments}
In this section, we demonstrate the performance of our proposed \name\ algorithm \footnote{\url{https://github.com/wulongfeng/CaliRare}} in terms of confidence calibration, rare category characterization, and parameter sensitivity analysis.

\begin{table}[htp]
\caption{Statistics of real-world graph benchmarks.}
\small
\centering
\label{tab: statistics}
\scalebox{0.9}{
 \begin{tabular}{cccccc}
\hline
\textbf{Types}& \textbf{Datasets}& \textbf{Nodes}& \textbf{Edges}& \textbf{Features}& \textbf{Minority Class}\\
\hline 
\multirow{3}{*}{Citation}
        &Cora  & 2708  & 10556   & 1433   & 12.96\% \\
        &CiteSeer  & 3327  & 9104   & 3703   & 15.72\% \\
         &PubMed  & 19717  & 88648   & 500   & 20.81\% \\
         &DBLP  & 17716  & 105734   & 1639  &  11.19\% \\
{Social}
        &FaceBook  &22470  &342004   &128 & 14.81\%\\
\hline
\end{tabular}}
\end{table}

\begin{table*}[!t]
    \caption{ Confidence Calibration evaluation on five datasets with various label rates for each class.} 
    \label{tab: calibration result}
    \centering
    \scalebox{0.95}{
    \begin{tabular}{ccccccccccccccccc}
        \hline
        \multirow{2}{*}{\textbf{DataSets}} &
        \textbf{LR\_C} & 
        \multicolumn{5}{c}{\textbf{20}} &
        \multicolumn{5}{c}{\textbf{30}} &
        \multicolumn{5}{c}{\textbf{40}} \\
        \hhline{~-~----~----~----}
        &\textbf{Metrics} & 
        & Recall & $F_1$ & ACE & \multicolumn{1}{c}{M\_ACE} & 
        & Recall & $F_1$ & ACE & \multicolumn{1}{c}{M\_ACE} & 
        & Recall & $F_1$ & ACE & \multicolumn{1}{c}{M\_ACE} \\
        \hline
        \multirow{7}{*}{Cora} 
        &\textbf{Uncal} && 71.54 & 78.89 & 0.1515 & 0.1104 & & 71.54 & 76.98 & 0.1544 & 0.1091 && 79.23 & 78.66 & 0.1606 & 0.1096 \\
        &\textbf{TS} && 72.31 & 81.82 & 0.2130 & 0.1235 & & 94.62 & 75.06 & 0.1230 & 0.0925 && 90.00 & 78.78 & \textbf{0.0930} & \textbf{0.0760} \\
        &\textbf{MS} && 53.85 & 78.89 & 0.326 & 0.1955 & & 53.08 & 78.38 & 0.3450 & 0.2095 && 63.08 & 82.70 & 0.3290 & 0.1930 \\
        &\textbf{LS} && \textbf{84.62} & 80.98 & 0.1507 & 0.1258 & & 80.00 & \textbf{82.70} & 0.1237 & 0.1166 && 86.92 & 81.81 & 0.1136 & 0.1048 \\
        &\textbf{MixUp} && \textbf{84.62} & 76.76 & 0.1414 & 0.1384 & & \textbf{97.69} & 46.78 & 0.1464 & 0.3495 && \textbf{96.92} & 52.01 & 0.1332 & 0.3438 \\
        &\textbf{CaGCN} && 50.00 & 77.68 & 0.2994 & 0.1594 & & 50.77 & 78.55 & 0.2921 & 0.1559 && 60.00 & 81.82 & 0.2298 & 0.1270 \\ 
        \hhline{~-~----~----~----}
        &\textbf{Ours} && \textbf{84.62} & \textbf{82.10} & \textbf{0.1263} & \textbf{0.0894}  & & 80.77 & 81.05 & \textbf{0.0958} & \textbf{0.0731} & & 86.15 & \textbf{83.00} & 0.1049 & 0.0816 \\
        \hline
        \multirow{7}{*}{PubMed} 
        &\textbf{Uncal} && 85.00 & 83.55 & 0.1217 & 0.0954 & & 83.89 & 83.31 & 0.0953 & 0.0847 && 87.22 & 82.82 & 0.0849 & 0.0694 \\
        &\textbf{TS} && 82.78 & 84.57 & 0.1200 & 0.0840 & & 84.44 & 85.26 & 0.0900 & 0.0700 && 88.89 & 83.59 & 0.0830 & 0.0755 \\
        &\textbf{MS} && 76.67 & \textbf{86.90} & 0.1650 & 0.1055 & & 78.89 & \textbf{87.27} & 0.1320 & 0.0875 && 77.78 & 85.76 & 0.1460 & 0.1000 \\
        &\textbf{LS} && 87.22 & 84.40 & 0.1100 & 0.1116 & & 90.00 & 83.58 & 0.1151 & 0.1143 && 93.89 & 81.87 & 0.1172 & 0.1118 \\
        &\textbf{MixUp} && \textbf{98.33} & 72.93 & 0.1145 & 0.1266 & & \textbf{95.56} & 77.74 & 0.1036 & 0.1230 && \textbf{98.89} & 78.07 & 0.1088 & 0.1117 \\
        &\textbf{CaGCN} && 73.89 & 86.31 & 0.1418 & 0.0760 & & 74.44 & 86.08 & 0.1504 & 0.0795 && 77.78 & \textbf{87.01} & 0.1620 & 0.0851 \\ 
        \hhline{~-~----~----~----}
        &\textbf{Ours} && 86.67 & 84.09 & \textbf{0.0688} & \textbf{0.0543} & & 88.33 & 84.38 & \textbf{0.0765} & \textbf{0.0565} & & 84.44 & 84.11 & \textbf{0.0701} & \textbf{0.0572} \\
        \hline
       \multirow{7}{*}{CiteSeer} 
        &\textbf{Uncal} && 69.37 & 84.10 & 0.1547 & 0.1281 & & 71.88 & 85.03 & 0.1470 & 0.1190 && 72.50 & 83.63 & 0.1514 & 0.1248 \\
        &\textbf{TS} && 68.13 & 84.54 & 0.2820 & 0.1730 & & 68.13 & 84.69 & 0.2190 & 0.1365 && 68.75 & 84.93 & 0.2080 & 0.1155 \\
        &\textbf{MS} && 62.50 & 83.87 & 0.3420 & 0.2120 & & 63.12 & 83.33 & 0.3340 & 0.2035 && 63.75 & 84.22 & 0.2220 & 0.1485 \\
        &\textbf{LS} && 68.75 & 84.16 & 0.1359 & 0.1121 & & 71.25 & 84.80 & 0.1551 & 0.1197 && 71.25 & 85.26 & 0.1422 & 0.1194 \\
        &\textbf{MixUp} && \textbf{81.25} & 84.00 & 0.1249 & 0.1427 & & \textbf{81.87} & 83.94 & 0.1085 & 0.1171 && \textbf{83.75} & 82.83 & 0.1261 & 0.1270 \\
        &\textbf{CaGCN} && 56.25 & 81.67 & 0.2842 & 0.1606 & & 64.38 & 83.68 & 0.2374 & 0.1299 && 58.75 & 82.44 & 0.2642 & 0.1498 \\ 
        \hhline{~-~----~----~----}
        &\textbf{Ours} && 75.00 & \textbf{85.72} & \textbf{0.1034} & \textbf{0.0957} & & 73.12 & \textbf{86.27} & \textbf{0.0961} & \textbf{0.0933} & & 75.00 & \textbf{86.79} & \textbf{0.1223} & \textbf{0.1037} \\
        \hline
        \multirow{7}{*}{DBLP} 
        &\textbf{Uncal} && 71.82 & 85.15 & 0.1907 & 0.1413 & & 73.64 & 81.70 & 0.1649 & 0.1254 && 78.18 & 81.35 & 0.1317 & 0.1031 \\
        &\textbf{TS} && 71.82 & 87.44 & 0.2440 & 0.1350 & & 79.09 & 87.43 & 0.1960 & 0.1135 && 83.64 & 86.93 & 0.2030 & 0.1245 \\
        &\textbf{MS} && 61.82 & 84.74 & 0.2430 & 0.1520 & & 63.64 & 84.37 & 0.2330 & 0.1440 && 60.00 & 82.51 & 0.3380 & 0.2025 \\
        &\textbf{LS} && 78.18 & \textbf{89.01} & 0.1457 & 0.1406 & & 89.09 & \textbf{87.47} & 0.1713 & 0.1530 && 89.09 & 84.49 & 0.1511 & 0.1381 \\
        &\textbf{MixUp} && \textbf{90.91} & 67.79 & 0.1046 & 0.1042 & & \textbf{98.18} & 57.96 & 0.1061 & 0.0930 && \textbf{94.55} & 68.24 & 0.0953 & 0.0921 \\
        &\textbf{CaGCN} && 55.45 & 82.51 & 0.2212 & 0.1194 & & 63.64 & 84.37 & 0.2181 & 0.1153 && 60.91 & 82.87 & 0.2292 & 0.1219 \\ 
        \hhline{~-~----~----~----}
        &\textbf{Ours} && 84.45 & 86.93 & \textbf{0.1026} & \textbf{0.0902} & & 82.73 & 86.27 & \textbf{0.1009} & \textbf{0.0819} & & 85.45 & \textbf{87.88} & \textbf{0.0964} & \textbf{0.0813} \\
        \hline
        \multirow{7}{*}{FaceBook} 
        &\textbf{Uncal} && 74.82 & 62.47 & 0.1180 & 0.1096 & & 69.78 & 63.73 & 0.1310 & 0.1006 && 89.93 & 64.23 & 0.1350 & 0.1091 \\
        &\textbf{TS} && 79.14 & 64.71 & 0.1660 & 0.1200 & & 74.10 & 64.75 & 0.1550 & 0.1215 && 91.37 & 65.18 & 0.1600 & 0.1190 \\
        &\textbf{MS} && 0.00 & 46.27 & 0.7240 & 0.4215 & & 5.04 & 46.27 & 0.7330 & 0.4255 && 5.76 & 51.78 & 0.5980 & 0.3555 \\
        &\textbf{LS} && 86.33 & 63.62 & 0.1337 & 0.1054 & &77.70 & 61.62 & 0.1127 & 0.0964 && 94.24 & 59.28 & 0.1377 & 0.1248 \\
        &\textbf{MixUp} && \textbf{96.40} & 47.24 & 0.1616 & 0.2028 & & \textbf{84.89} & 62.85 & 0.1010 & 0.1171 && \textbf{97.12} & 55.51 & 0.1122 & 0.1436 \\
        &\textbf{CaGCN} && 42.45 & 68.85 & 0.4340 & 0.2358 & &43.17 & 72.00 & 0.3869 & 0.2116  && 62.59 & \textbf{81.31} & 0.2164 & 0.1203 \\ 
        \hhline{~-~----~----~----}
        &\textbf{Ours} && 79.14 & \textbf{73.14} & \textbf{0.0843} & \textbf{0.0583} & & 75.54 & \textbf{75.64} & \textbf{0.0962} & \textbf{0.0788} & & 80.58 & 77.93 & \textbf{0.1109} & \textbf{0.0833} \\
        \hline
    \end{tabular}}
\end{table*}

\subsection{Experiment Setup}
\textbf{Dataset:} We utilize five commonly used real-world graph benchmarks: Cora~\cite{sen2008collective}, CiterSeer~\cite{sen2008collective}, PubMed~\cite{sen2008collective}, DBLP~\cite{fu2020magnn}, and FaceBook~\cite{rozemberczki2021multi}. The statistics of these datasets are summarized in Table~\ref{tab: statistics}, and the last column is the ratio of the minority class. Specifically, to further augment the imbalance of rare category problems, we choose one class of the dataset as the minority class, and the rest are combined to form the majority class. To provide more results, we select three label rates for the training set, specifically, 20, 30, and 40 nodes for each class before combination.

\begin{table*}[!th]
    \caption{Rare category characterization evaluation on five datasets with various label rates for each class.}
    \label{tab: classification result}
    \centering
    \scalebox{0.99}{
    \begin{tabular}{cccccccccccccc}
        \hline
        \multirow{2}{*}{\textbf{DataSets}} &
        \textbf{LR\_C} & 
        \multicolumn{4}{c}{\textbf{20}} &
        \multicolumn{4}{c}{\textbf{30}} &
        \multicolumn{4}{c}{\textbf{40}} \\
        \hhline{~-~---~---~---}
        &\textbf{Metrics} & 
        & Acc & Recall & $F_1$ & 
        & Acc & Recall & $F_1$ & 
        & Acc & Recall & $F_1$ \\
        \hline
        \multirow{4}{*}{Cora} 
        &\textbf{GIN-CS} && 82.70 & 43.08 & 39.30 & & 79.90 & 46.15 & 37.38 & & 76.80 & 66.15 & 42.57\\
        &\textbf{GAT-CS} && \textbf{90.50} & 62.31 & 63.04 & & 89.60 & 55.38 & 58.06 && 80.20 & 82.31 & 51.94\\
        &\textbf{GCN-CS} && 89.50 & 71.54 & 78.89 & & 88.10 & 71.54 & 76.98 && 88.50 & 79.23 & 78.66\\
        \hhline{~-~---~---~---}
        &\textbf{Ours} && \textbf{90.50} & \textbf{84.62} & \textbf{82.10} & & \textbf{90.10} & \textbf{80.77} & \textbf{81.05} & & \textbf{91.00} & \textbf{86.15} & \textbf{83.00}\\
        \hline
        \multirow{4}{*}{PudMed} 
        &\textbf{GIN-CS} && 80.00 & 36.11 & 39.39 & & 81.90 & 32.78 & 39.46 & & 76.80 & 47.22 & 42.29\\
        &\textbf{GAT-CS} && 87.80 & 78.89 & 69.95 & & 87.70 & \textbf{88.89} & 72.23 && 88.00 & 77.78 & 70.00\\
        &\textbf{GCN-CS} && 89.20 & 85.00 & 83.55 & & 89.10 & 83.89 & 83.31 && 88.40 & \textbf{87.22} & 82.82\\
        \hhline{~-~---~---~---}
        &\textbf{Ours} && \textbf{89.50} & \textbf{86.67} & \textbf{84.09} & & \textbf{89.60} & 88.33 & \textbf{84.38} & & \textbf{89.70} & 84.44 & \textbf{84.11}\\
        \hline
        \multirow{4}{*}{CiteSeer} 
        &\textbf{GIN-CS} && 75.80 & 50.00 & 39.80 & & 71.40 & 53.12 & 37.28 & & 71.70 & 68.13 & 43.51\\
        &\textbf{GAT-CS} && 92.30 & 69.37 & 74.25 & & 91.30 & \textbf{76.88} & 73.87 && 92.10 & \textbf{75.00} & 75.24\\
        &\textbf{GCN-CS} && 91.80 & 69.37 & 84.10 & & 92.20 & 71.88 & 85.03 && 91.20 & 72.50 & 83.63\\
        \hhline{~-~---~---~---}
        &\textbf{Ours} && \textbf{92.40} & \textbf{75.00} & \textbf{85.72} & & \textbf{92.90} & 73.12 & \textbf{86.27} & & \textbf{93.10} & \textbf{75.00} & \textbf{86.79}\\
        \hline
        \multirow{4}{*}{DBLP} 
        &\textbf{GIN-CS} && 81.00 & 40.91 & 32.17 & & 76.40 & 43.64 & 28.92 & & 88.20 & 20.00 & 27.16\\
        &\textbf{GAT-CS} && 94.30 & 81.82 & 97.95 & & 93.60 & 64.55 & 68.93 && 93.20 & 67.27 & 68.52\\
        &\textbf{GCN-CS} && 94.30 & 71.82 & 85.15 & & 92.30 & 73.64 & 81.70 && 91.70 & 78.18 & 81.35\\
        \hhline{~-~---~---~---}
        &\textbf{Ours} && \textbf{94.40} & \textbf{85.45} & \textbf{86.93} && \textbf{94.20} & \textbf{82.73} & \textbf{86.27} && \textbf{94.90} & \textbf{85.45} & \textbf{87.88} \\
        \hline
        \multirow{4}{*}{FaceBook} 
        &\textbf{GIN-CS} && 79.30 & 63.31 & 45.95 & & 75.50 & 74.10 & 45.68 & & 79.60 & 67.63 & 47.96\\
        &\textbf{GAT-CS} && 81.70 & 74.10 & 52.96 & & \textbf{88.70} & 59.71 & 59.50 && 83.50 & 76.26 & 56.23\\
        &\textbf{GCN-CS} && 72.50 & 74.82 & 62.47 & & 74.90 & 69.78 & 63.73 && 72.20 & \textbf{89.93} & 64.23\\
        \hhline{~-~---~---~---}
        &\textbf{Ours} && \textbf{83.20} & \textbf{79.14} & \textbf{73.14} & & 85.90 & \textbf{75.54} & \textbf{75.64} & & \textbf{87.20} & 80.58 & \textbf{77.93} \\
        \hline
    \end{tabular}}
\end{table*}

\subsection{Calibration Evaluation}

\textbf{Baselines for Calibration.} We conduct comparison experiments between our method and the following state-of-the-art calibration approaches. We follow the official implementation and use the default setting on hyperparameters (if needed) for these baselines.
\begin{itemize}
\item {\bfseries Temperature Scaling (TS)}~\cite{guo2017calibration}: Temperature scaling is a simple extension of Platt scaling. It learns a scalar parameter $t$ as the temperature of the predictive distribution and smoothes the predicted probability by the temperature.
\item {\bfseries Matrix Scaling (MS)}~\cite{guo2017calibration}: Matrix scaling is another extension of Platt scaling. It replaces the scalar parameter with a matrix and outputs the calibrated predictive probability.
\item {\bfseries Label Smoothing (LS)}~\cite{szegedy2016rethinking}: Label smoothing replaces one-hot encoded labels with a mixture of the label and a uniform distribution. ~\cite{muller2019does} demonstrates that it can implicitly calibrate the model's predictions as it can prevent the model from overfitting/over-confidence.
\item {\bfseries Mixup}~\cite{wang2021mixup}: As a popular data augmentation technique, Mixup takes combinations of pairs of training samples and their labels. It has been shown that Mixup can significantly improve confidence calibration across diverse applications.
\item {\bfseries CaGCN}~\cite{wang2021confident}: CaGCN is a method for confidence calibration in GNNs. It learns a unique transformation from the logits of GNNs to calibrate the confidence of each node. 
\end{itemize}

\noindent\textbf{Experimental settings.}
We evaluate the performance of all methods by three standard measurements: recall, Macro-F1, and ACE. Compared with the majority class, the ratio of the minority class is very small. Thus, recall reflects the performance of models on the minority class more precisely. Macro-F1 is the unweighted mean of F1 for each class, and will not be affected by data imbalance. ACE is a popular metric for calibration, specifically, it refers to the ACE of the minority class here. Furthermore, to have a more comprehensive understanding of the results, we propose a new metric: Macro-ACE, which is the unweighted mean of ACE for each class. For Cora/PubMed/CiteSeer, we adopt the default training set (20 nodes per class), validation set (500 nodes), and test set (1000 nodes) provided by torch\_geometric, and randomly chose nodes in the same setting for the rest of datasets. We also randomly add nodes to the training set for different label rates. Additionally, as presented in Table~\ref{tab: statistics}, the minority class accounts for a small portion of the datasets, so the cost-sensitive loss is utilized to increase the importance of the minority class for all methods, and the weights are inversely proportional to their percentage.
The base uncalibrated model is a GCN model with parameters suggested by ~\cite{kipf2016semi, wang2021confident}. For our method, we also adopt a two-layer GCN with the same hyper-parameters of GCN, and set the coverage parameter $\alpha=0.9$ and the trade-off hyper-parameter $\lambda=0.1$ for all datasets. 

\noindent\textbf{Results.}
The calibration evaluation results are reported in Table ~\ref{tab: calibration result}. In general, we have those observations: (1) Our method is better than Uncal across almost all metrics, demonstrating that our method can calibrate the confidence without compromising its accuracy and recall. (2) Compared with other methods, MixUp shows good performance regarding the recall rather than Macro-F1.  (3) In contrast to Mixup, MS focuses more on accuracy than recall and will get overfitting on some datasets. (4) CaGCN performs significantly better on the majority class than the minority class, that's why it has larger ACE and normal Macro-ACE.

\subsection{Rare Category Characterization}
\textbf{Baselines for Classification.} We conduct comparison experiments between our method and the following approaches. Specifically, we consider popular GNNs with cost-sensitive loss as our baseline methods in the task of rare category characterization. We use the default setting on hyperparameters (if needed) for these baselines.
\begin{itemize}
\item {\bfseries Cost-Sensitive Graph Convolutional Network (GCN-CS)}: GCN~\cite{kipf2016semi} is a type of convolutional neural network. It applies convolution on graphs and takes advantage of the structural information.
\item {\bfseries Cost-Sensitive Graph Attention Network (GAT-CS)}~\cite{velivckovic2017graph}: GAT is a popular method in GNNs. It expands the basic aggregation function of the GCN and assigns different importance to each edge through the attention mechanism. 
\item {\bfseries Cost-Sensitive Graph Isomorphism Network (GIN-CS)}: GIN ~\cite{xu2018powerful} is another method in GNNs that aims to achieve the ability as the Weisfeiler-Lehman graph isomorphism test.
\end{itemize}

\noindent\textbf{Experimental settings.}
We evaluate the performance of all methods by three standard measurements for classification: accuracy, recall, and Macro-F1. It shares the same experimental setting with the experiments of calibration.

\noindent\textbf{Results.}
The classification evaluation results are reported in Table ~\ref{tab: classification result}. In general, we have the following observations: (1) our method achieves better performance than the baseline methods across all metrics. For example, compared to our best competitor GAT-CS, we achieve a 28.77\% improvement in recall and 43.21\% improvement in Macro-F1 on the dataset of Cora. (2) More label rates for each class will get better performance, which is consistent with our understanding.
(3) In some cases, GAT-CS have a better recall. For example, for Citeseer with 30 label rates, the recall of GAT-CS is larger than ours, but Macro-F1 is lower than ours, that's because it misclassifies majority class examples as minority class examples.

\subsection{Parameter Sensitivity Analysis}
\begin{figure}[!tb]
\centering
\includegraphics[width=.46\textwidth]{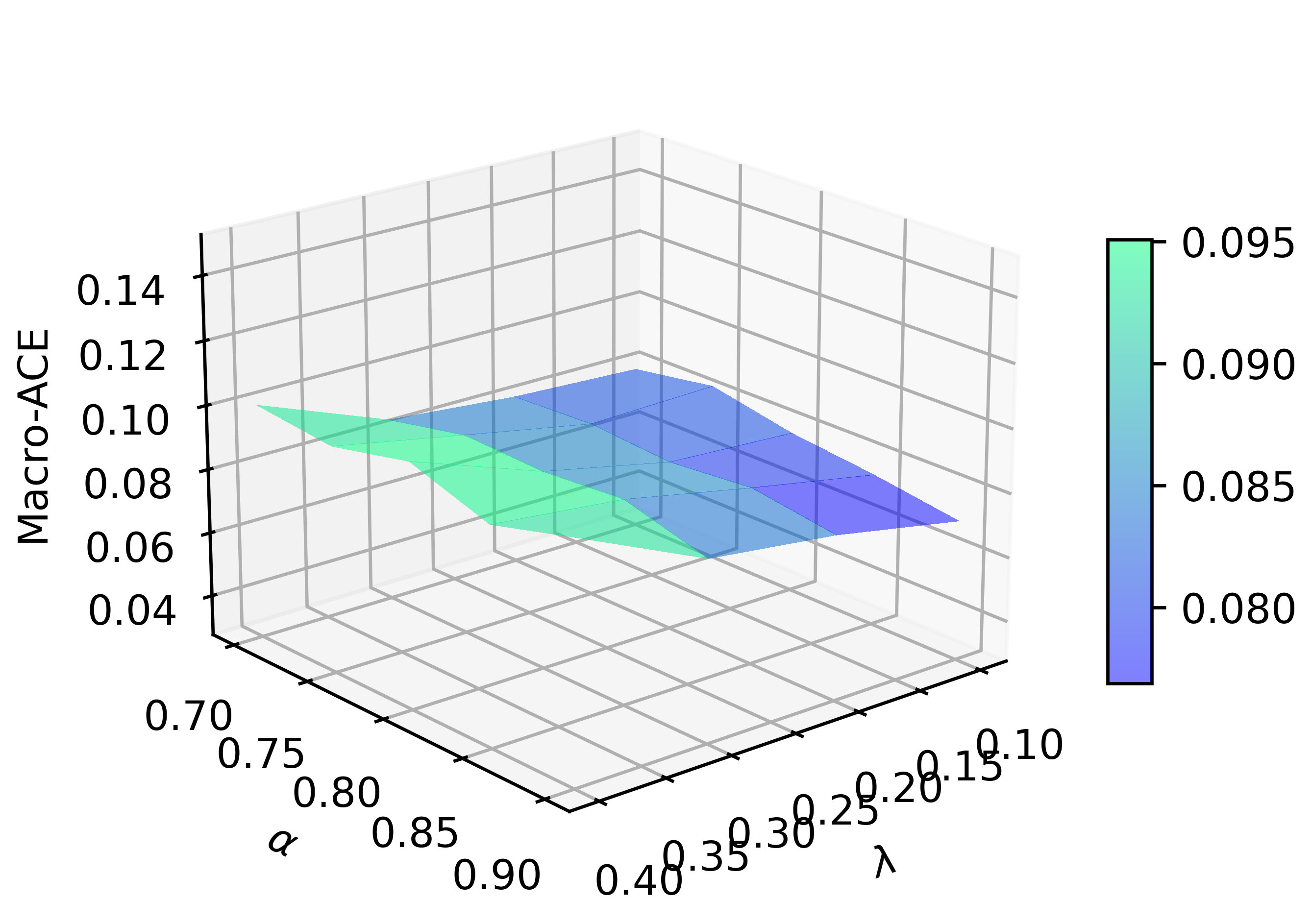}
\caption{Parameter study on the coverage parameter $\alpha$ and hyper-parameter $\lambda$.}
\label{fig: parameter sensitivity}
\vspace{-5mm}
\end{figure}

We study the parameter sensitivity of our proposed method \name\ with respect to the coverage parameter $\alpha$ and the trade-off parameter $\lambda$. Here, given $\alpha \in [0.7, 0.75, 0.8, 0.85, 0.9]$ and $\lambda \in [0.1, 0.2, 0.3, 0.4]$, we test these parameters on Cora for the label rate of each class is 20 in terms of Macro-ACE. The results are shown in Figure~\ref{fig: parameter sensitivity}, and we can get the following observations: (1) Compared with $\alpha$, Macro-ACE is more likely to be affected by $\lambda$; (2) As the value of $\lambda$ increases, the importance of calibration also increases, which degrade the model's accuracy and affect calibration.

\section{Related Work}
In this section, we briefly review the recent advances in rare category characterization and confidence calibration.

\textbf{Rare Category Characterization.} Pelleg et al. first formalize the problem of Rare category characterization~\cite{pelleg2004active}. In contrast to imbalance classification~\cite{sun2009classification, chawla2002smote}, which seeks to improve the overall performance, and outlier detection~\cite{hodge2004survey}, which looks for irregular patterns, rare category characterization investigates the compactness of the minority class and characterizes them from the highly-skewed datasets. Extensive work has been done for rare category characterization. GRADE ~\cite{he2008graph} is proposed by He et al., it detects rare categories from regions where probability density changes the most. MUVIR~\cite{zhou2015muvir} exploits the relationship between multiple views to characterize rare categories, and every single view is built based on existing techniques for rare category detection. Some sampling-based methods are also introduced, such as SMOTE~\cite{chawla2002smote}, which adopts the technique of a combination of over-sampling the minority classes and under-sampling the majority class to achieve better performance. Other representation learning methods are also proposed, such as SPARC~\cite{zhou2018sparc} etc. However, these methods are mainly designed to improve accuracy, little attention has been paid to the calibration of rare categories.

\textbf{Confidence Calibration.} Confidence calibration has been well-studied in various fields recently. Niculescu-Mizil et al. propose to utilize reliability diagrams, and useful visual tools to explore calibration of deep neural network~\cite{niculescu2005predicting}. Guo et al. further introduce a more convenient metric, the expected calibration error (ECE), as it is a scalar summary statistic of calibration~\cite{guo2017calibration}, which is commonly used in the field of calibration. They also found that modern neural networks are poorly calibrated and investigated factors that influence calibration. Platt scaling ~\cite{platt1999probabilistic} is the most popular post-processing parametric approach. It learns two scalar parameters on the validation set and outputs the calibrated probability based on them. Temperature scaling ~\cite{guo2017calibration} is a simple extension of Platt scaling, it only involves a single scalar, namely temperature for all classes. Matrix scaling and vector scaling are another two extensions of Platt scaling to higher-dimension. Isotonic regression~\cite{zadrozny2002transforming} is the most popular non-parametric method that finds the piecewise constant function to calibrate probabilities. Wang et al. explore the confidence calibration in GNNs~\cite{wang2021confident}. However, little work has been done in the scenario of rare category characterization. Additionally, the shortcomings of ECE have been demonstrated in many works ~\cite{nixon2019measuring, vaicenavicius2019evaluating, zhang2020mix, ashukha2020pitfalls}, such as fixed calibration ranges, pathologies in static binning schemes, etc. Other metrics, such as adaptive calibration error (ACE) are introduced to address these issues, which are also used in our paper. We further propose the EICE, calibrating the model without compromising its accuracy.

\section{Conclusion}
Rare categories are of great importance to a variety of high-impact applications, such as financial fraud detection, rare disease diagnosis, etc. However, most of the existing work is designed to improve the accuracy of rare category analysis, while little attention has been paid to enhancing its reliability. 
In this paper, we propose EICE, which leverages individual uncertainty to approximate its accuracy and measures the calibration error at the level of instances. We further design a regularization term based on EICE, jointly learn the characterizations of rare categories and calibrate the confidence. Extensive results show that our method achieves significant improvements for rare categories analysis by comparing with five popular baseline methods for calibration and three popular baseline methods for classification.

\begin{acks}
This work is supported by Virginia Tech, Cisco, Deloitte, Commonwealth Cyber Initiative, and 4-VA. The views and conclusions are those of the authors and should not be interpreted as representing the official policies of the funding agencies or the government.
\end{acks}

\bibliographystyle{ACM-Reference-Format}
\bibliography{Calibration on Rare Category}


\end{document}